\documentclass[]{companypaper}

\partnerlogos{\includegraphics[height=10mm]{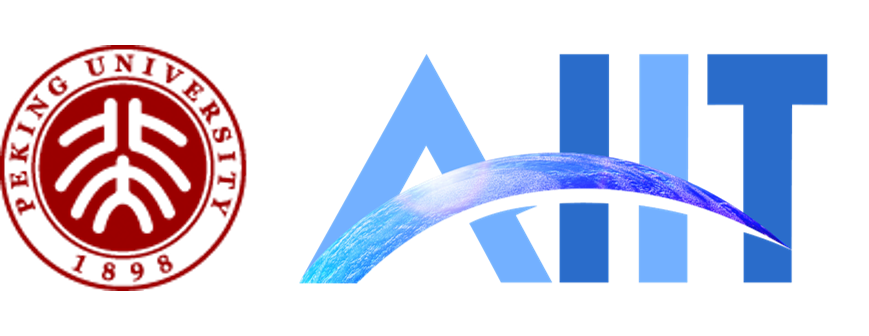}}  
\reportseries{MEMO · TECHNICAL REPORT}   

\usepackage[toc,page,header]{appendix}
\usepackage{microtype}
\usepackage{graphicx}
\usepackage{subcaption}
\usepackage{booktabs}
\usepackage{hyperref}
\usepackage{bm}
\usepackage{amsmath}
\usepackage{amssymb}
\usepackage{mathtools}
\usepackage{amsthm}
\theoremstyle{plain}

\theoremstyle{definition}

\theoremstyle{remark}

\usepackage{xspace}
\usepackage{enumitem}
\usepackage{wrapfig}
\usepackage{multirow}
\usepackage{tabularx}
\usepackage{colortbl}
\usepackage{float}

\definecolor{methodviolet}{RGB}{109,40,217}
\definecolor{benchteal}{RGB}{13,148,136}

\newcommand{\figref}[1]{\hyperref[#1]{Fig.~\ref*{#1}}}
\newcommand{\tabref}[1]{\hyperref[#1]{Table~\ref*{#1}}}
\newcommand{\secref}[1]{\hyperref[#1]{Section~\ref*{#1}}}
\newcommand{\eqnref}[1]{\hyperref[#1]{Eq.~\ref*{#1}}}

\title{RoboTwin-Phys: Do WAMs and VLAs Understand the Physical World?}

\author[1,2,3]{Jiaqi Zhang}
\author[1,\dagger]{Feng Ye}
\author[1]{Mingjia Yang}
\author[1]{Zhihong Chen}
\author[1]{Mingkang Xiang}
\author[1]{Xinglin Yao}
\author[1]{Yanbin Li}
\author[1]{Siwei Ma}
\author[1, \ddagger]{Chuanmin Jia}

\affiliation[1]{Peking University}
\affiliation[2]{Advanced Institute of Information Technology, Peking University}
\affiliation[3]{Memo}

\contribution[\dagger]{Project Lead}
\contribution[\ddagger]{Corresponding Author}

\date{\today}
\abstract{%
Physical-condition diversity is largely missing from current benchmarks for robot manipulation. While large-scale simulation benchmarks increasingly incorporate variations in object appearance, scene layout, and visual observations, they typically keep the underlying physical parameters fixed. As a result, important sources of real-world variability, such as changes in mass, friction, and joint dynamics, remain largely untested.
We introduce \textbf{RoboTwin-Phys}, a physics-diverse benchmark that treats physical-condition diversity as an explicit dimension of robot manipulation evaluation. The benchmark continuously varies \textbf{13 physical attributes} within physically plausible ranges, providing a unified setting for evaluating policies across diverse physical operating conditions. We further release more than 5,000 expert demonstrations with ground-truth physical parameters, enabling physical-attribute estimation, condition-aware modeling, and physics-conditioned policy training.
Evaluations of representative WAMs and VLAs reveal a substantial robustness gap: models that remain effective under existing visual and layout randomization can degrade markedly under changes in physical conditions. RoboTwin-Phys provides the benchmark, data, and evaluation protocol needed to systematically measure and improve robustness to physical-condition diversity in robot manipulation.
}

\begin{document}
\maketitle

\vspace{-5mm}
\noindent
\begin{minipage}{\textwidth}
\centering
\includegraphics[width=0.98\textwidth]{figures/teaser.jpg}
\refstepcounter{figure}\label{fig:teaser}
\par\vspace{0.2em}
{\footnotesize\textbf{Figure \thefigure:} \textbf{Overview of Robotwin-Phys.}}
\end{minipage}
\vspace{0.3em}

\section{Introduction}

Vision-Language-Action (VLA) models and World-Action Models (WAMs) are pushing robot learning from isolated skills toward general-purpose manipulation. Along this trend, large-scale bimanual simulation benchmarks such as RoboTwin-2.0~\cite{chen2025robotwin} provide standardized task suites, expert demonstrations, and evaluation protocols, enabling reproducible training and systematic comparison across models.

However, an important source of environmental variability remains largely unmeasured: \textbf{physical conditions}. Existing manipulation benchmarks increasingly diversify what a robot sees through changes in object appearance, scene layout, object pose, lighting, and camera viewpoints, but the underlying physical parameters are often kept at fixed nominal values. In the real world, however, such parameters vary naturally across otherwise similar operating conditions. Object mass changes with loading; friction depends on material and surface wear; the actual center of mass (CoM) can deviate from its nominal model due to manufacturing tolerances or object contents; and joint damping can change with mechanical wear. These factors directly affect contact dynamics and the outcome of manipulation, yet they are rarely treated as an explicit source of benchmark diversity. Consequently, performance under visual and layout variation does not by itself reveal how robust a model is to changes in the physical properties that govern action execution.

This gap has three direct consequences. \textbf{First, no standardized evaluation}: existing benchmarks provide no systematic protocol for measuring how manipulation performance changes as physical conditions vary. \textbf{Second, limited training resources}: there is no large-scale dataset that deliberately covers diverse physical conditions across a broad set of manipulation tasks, limiting the development of condition-aware models and policies. \textbf{Third, missing physical ground truth}: without the actual physical parameters associated with each sample, it is difficult to train, analyze, or audit modules that are intended to infer or utilize physical conditions. Existing studies of physical factors are predominantly based on isolated factors and small-scale ablations, making it difficult to characterize joint variation across factors or turn these observations into a reusable benchmark and data resource.

We present \textbf{RoboTwin-Phys}, a systematic benchmark for physical-condition diversity built on the 50-task bimanual platform of RoboTwin-2.0. Rather than introducing new manipulation tasks, RoboTwin-Phys augments the existing task suite with a dedicated physical variation dimension. Specifically, \textbf{13 physical attributes} are continuously sampled at the episode level within physically plausible ranges, allowing each episode to instantiate a different physical operating condition. 

Beyond evaluation, we construct a training set of more than \textbf{5,000 expert demonstrations} covering the resulting distribution of physical conditions. The data remain fully compatible with the official RoboTwin format, while each sample is additionally annotated with the 13-dimensional ground-truth physical parameters in effect during the episode. At evaluation time, the physical attributes are sampled during environment initialization to instantiate the episode-specific operating condition. Expert planning is performed in advance to verify feasibility under the sampled setting, ensuring that the benchmark evaluates models on physically realizable conditions rather than arbitrary parameter perturbations. These design choices make RoboTwin-Phys suitable not only for robustness evaluation, but also for physical-attribute estimation, condition-aware modeling, and physics-conditioned policy learning.

Using this benchmark, we evaluate representative WAMs and VLAs under diverse physical conditions. Our experiments reveal a consistent robustness gap: models that remain effective under existing visual and layout randomization can experience substantial degradation when the physical conditions are varied. This result shows that physical-condition diversity exposes a distinct failure dimension that is largely invisible to current benchmark protocols, motivating explicit modeling and evaluation of physical variability in robot learning.

The main contributions are summarized as follows:

\begin{enumerate}
    \item \textbf{RoboTwin-Phys benchmark.} We establish RoboTwin-Phys, a systematic benchmark for physical-condition diversity, introducing 13 continuously sampled physical attributes with task-specific calibrated ranges on RoboTwin~2.0.
    \item \textbf{Physics-annotated training set.} We release more than 5000 expert demonstrations, covering diverse physical conditions and annotated with 13-dimensional ground-truth physical parameters. The dataset directly supports physical-condition estimation, condition-aware representation learning, and physics-conditioned policy training. 
    \item \textbf{Evaluation of physical robustness.} We introduce a systematic evaluation protocol for WAMs and VLAs under physical-condition diversity and provide comparative experiments that characterize the resulting robustness gap across representative models. 
\end{enumerate}
    
\section{RoboTwin-Phys Benchmark}
\label{sec:benchmark}

\subsection{Benchmark Design}
\label{sec:design}

RoboTwin-Phys introduces \textbf{physical-condition diversity} as a new dimension of robot manipulation evaluation. Existing benchmarks have made substantial progress in diversifying visual appearance, scene layout, and object configuration, but the physical properties are typically fixed at nominal values. RoboTwin-Phys explicitly varies these physical conditions and evaluates whether a manipulation policy remains effective across the resulting operating environments.

The benchmark is built around three design principles. First, \textbf{physical validity}: each sampled condition should correspond to a physically realizable operating environment, rather than an arbitrary simulator perturbation. Second, \textbf{continuous diversity}: physical factors are varied continuously over meaningful ranges instead of being reduced to a few manually selected cases. Third, \textbf{composability}: physical-condition variation is treated as an independent evaluation dimension and can be studied either alone or together with existing visual and layout variation.

Under this design, each episode is instantiated with a complete set of physical conditions before execution begins. The sampled conditions remain fixed throughout the episode, providing a consistent physical environment for the policy. The benchmark allows multiple physical factors to vary simultaneously, exposing models to combinations of conditions that more closely reflect the diversity of real operating environments.

\begin{figure}[htb]
  \centering
  \footnotesize
  
    \includegraphics[width=\linewidth]{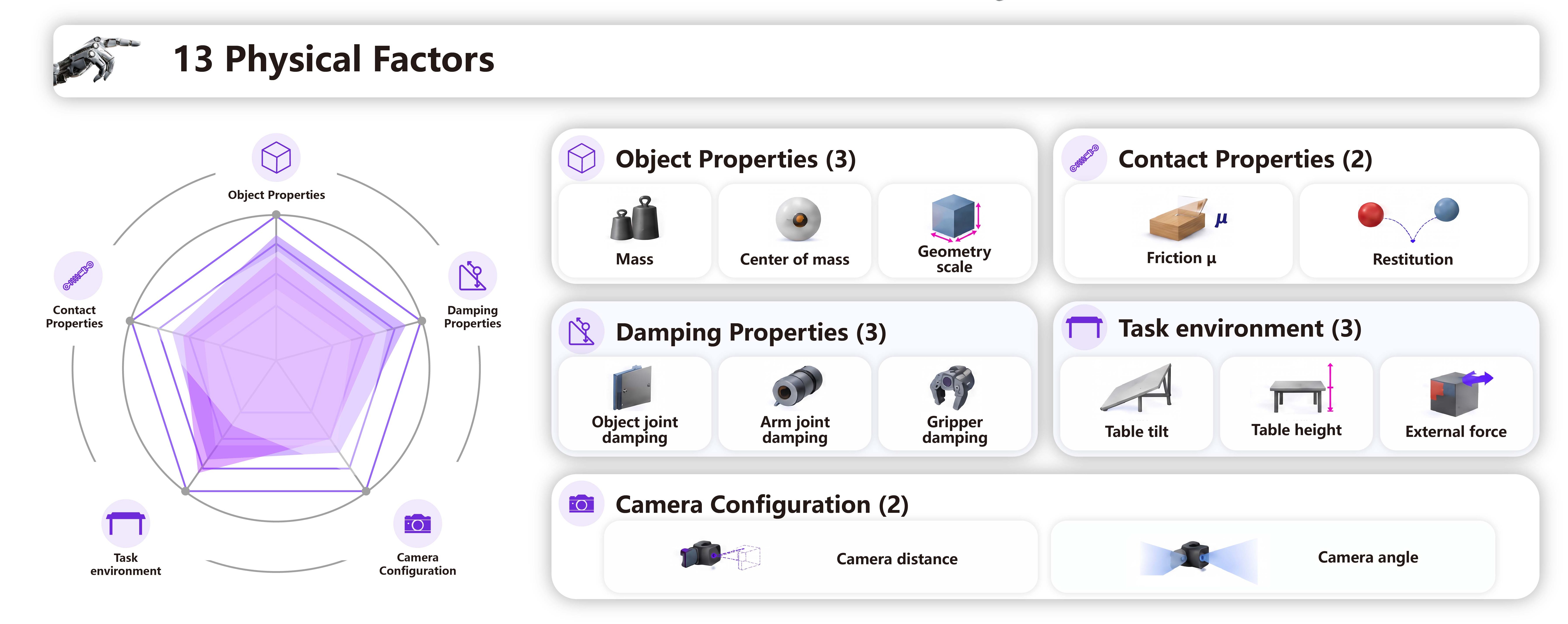}
  \caption{RoboTwin-Phys instantiates this design with 13 physical attributes, covering scene geometry, sensing configuration, mass, material properties, mass distribution, and joint characteristics. }
  \label{fig:attributes}
\end{figure}

As shown in Fig~\ref{fig:attributes}, RoboTwin-Phys instantiates this design with 13 physical attributes. Most tasks share a common physical-condition configuration, while physics-sensitive tasks use dedicated feasible domains to avoid physically implausible combinations. The complete attribute ranges are provided in Appendix~\ref{app:physical-config}.

\subsection{Physical-Condition Sampling}
\label{sec:physical-space}

The 13 physical attributes are sampled independently at the episode level from their designated continuous ranges. Sampling is performed during environment initialization, and the resulting values remain unchanged during the corresponding rollout. The episode seed determines the sampling process, making the instantiated physical condition reproducible.

The attributes cover complementary sources of physical variation, including object properties, contect properties, damping properties, task environments and camera configuration. Their ranges are chosen to represent plausible differences that may arise across objects, materials, mechanical states, or deployment environments.

Physical sampling is independent of the original visual and layout randomization. Consequently, RoboTwin-Phys supports three complementary evaluation settings: physical variation can be isolated to study physical robustness, visual variation can be retained as in the original benchmark, or both dimensions can be enabled simultaneously. This separation makes it possible to distinguish robustness to physical conditions from robustness to conventional visual and scene variation.

\subsection{Task-Aware Physical Validity}
\label{sec:validity}

A single global physical range is not necessarily appropriate for every manipulation task. The admissible range of a physical factor can depend on the geometry and operational constraints of the task. Applying a global range without considering these constraints may produce simulator configurations that are numerically valid but no longer represent plausible instances of the corresponding task.

Nine tasks are assigned dedicated configurations, determined through empirical calibration of their sensitive physical factors. During environment construction, each task is automatically routed to either the global configuration or its dedicated configuration.

This task-aware design is essential for ensuring that the benchmark measures robustness to physical diversity rather than robustness to artificially induced failure cases. The detailed task-specific configurations are provided in Appendix~\ref{app:task-config}.

\subsection{Feasibility and Benchmark Instances}
\label{sec:feasibility}

A sampled physical condition is retained as a benchmark instance only when the corresponding task remains executable under that condition. To verify this, the expert planner is run in advance on the instantiated environment. Episodes for which the expert cannot successfully complete the task are excluded from the evaluation pool.

This feasibility check ensures that benchmark failures reflect the evaluated model's ability to cope with physical variation, rather than the absence of a valid solution under the sampled environment. As a result, RoboTwin-Phys evaluates models over a collection of physically meaningful and task-feasible operating conditions.

The same principle is applied when generating the released expert demonstrations, ensuring consistency between the benchmark evaluation space and the accompanying training resource.

\section{Dataset}
\label{sec:dataset}

Together with the benchmark, we release \textbf{RoboTwin-Phys Training Set v1}, a large-scale collection of expert demonstrations sampled from the physical-condition space defined in Sec.~\ref{sec:physical-space}. The dataset is designed to make physical conditions directly observable as learning targets rather than hidden simulator states.

\subsection{Dataset Construction}
\label{sec:data-construction}

The training set contains more than 5{,}000 expert episodes spanning the physical-condition distribution. Demonstrations are generated with the established expert pipeline, and each retained episode is successfully executable under its instantiated physical condition. The collection therefore provides demonstrations from valid physical operating environments rather than arbitrary simulator configurations.

The physical sampling process is consistent with the benchmark definition. Most tasks use the global physical-condition configuration, while the physics-sensitive tasks are automatically routed to their dedicated configurations. The original visual and layout variation dimensions remain available during collection, and physical sampling is performed independently of these factors.

\subsection{Physical Ground Truth}
\label{sec:data-annotation}

Each demonstration is accompanied by the physical condition instantiated during its rollout. In particular, we provide the ground-truth values of the 13 physical attributes, enabling the physical state of the environment to be treated as an explicit supervision signal.
The detailed data format and organization are provided in Appendix~\ref{app:data-format}.

This explicit correspondence between observations, actions, and physical conditions enables several downstream settings, including physical-attribute estimation, physical-condition-aware representation learning, and policy training conditioned on physical attributes.

Overall, the dataset turns physical-condition diversity into an explicit and reusable learning resource, complementing the benchmark's role as an evaluation protocol.

\section{Experiments}
\label{sec:experiments}

We evaluate representative World-Action Models (WAMs) and Vision-Language-Action (VLA) models on RoboTwin-Phys to investigate their robustness to physical-condition diversity. Our experiments address three questions: 
(1) how current models perform when physical conditions vary, 
(2) whether the effect is consistent across different models, 
and (3) whether the resulting failures exhibit structured task-level patterns.

\subsection{Evaluation Setup}
\label{sec:eval-setup}

We evaluate representative World-Action Models (WAMs) and Vision-Language-Action (VLA) models using three environment protocols:

\begin{itemize}
    \item \textbf{Clean}: all benchmark randomization is disabled, corresponding to the nominal environment configuration.
    \item \textbf{Official Random}: the original visual and layout randomization dimensions are enabled, while physical parameters remain at their nominal values.
    \item \textbf{Physical Random}: the original randomization dimensions are retained and the 13 physical attributes are additionally sampled according to the RoboTwin-Phys protocol.
\end{itemize}

We evaluate five representative models, including Fast-WAM~\cite{yuan2026fastwam}, Motus~\cite{bi2025motus}, and FACT~\cite{peng2026fact}, $\boldsymbol{\pi}_{0.5}$~\cite{intelligence2025pi05} and Galaxea-VLA (G0.5)~\cite{liu2026g05auto}.

\subsection{Overall Results}
\label{sec:overall-results}

Table~\ref{tab:main-results} summarizes the performance on three settings.

\begin{table}[t]
    \centering
    \small
    \caption{Success rates (\%) of representative WAMs and VLAs on RoboTwin-2.0 settings and RoboTwin-Phys. Clean and Official Random are reported results from the corresponding public benchmark evaluations, while Physical Random is evaluated by us.}
    \label{tab:main-results}
    \begin{tabular}{@{}lccc@{}}
        \toprule
        Method & Clean & Official Random & Physical Random \\
        \midrule
        Fast-WAM~\cite{yuan2026fastwam} & 91.88 & 91.78 & \textbf{44.24} \\
        Motus~\cite{bi2025motus}    & 88.66 & 87.02 & \textbf{39.60} \\
        FACT~\cite{peng2026fact}     & 88.40 & 86.60 & \textbf{39.14} \\
        $\pi_{0.5}$~\cite{intelligence2025pi05} & 82.74 & 76.76 & \textbf{31.60} \\
        Galaxea-VLA~\cite{liu2026g05auto} & 93.70 & 92.80 & \textbf{37.82} \\
        \bottomrule
    \end{tabular}
\end{table}

A substantial performance gap emerges when physical conditions are varied. Under Physical Random, Fast-WAM, Motus, FACT, $\pi_{0.5}$, and Galaxea-VLA achieve success rates of 44.24\%, 39.60\%, 39.14\%, 31.60\%, and 37.82\%, respectively. All five evaluated models experience a large performance drop when moving from initial randomized environments to the physically diverse environments defined by RoboTwin-Phys.

In contrast, the differences between Clean and Official Random are comparatively small for the three WAMs: 0.10 points for Fast-WAM, 1.64 points for Motus, and 1.80 points for FACT. The corresponding decrease for $\pi_{0.5}$ is larger at 5.98 points, while Galaxea-VLA decreases by 0.90 points. 
This result exposes a robustness dimension that is distinct from conventional visual and layout randomization. Models that perform well when appearance and scene configuration change can still be highly sensitive to changes in the physical conditions governing interaction.

The effect is also observed across both WAMs and VLAs. Despite substantial differences in architecture and training paradigm, all five models exhibit pronounced degradation under Physical Random. This cross-family consistency suggests that the challenge revealed by RoboTwin-Phys is not specific to a particular model design.

\subsection{Task-Level Analysis}
\label{sec:task-analysis}

The aggregate results conceal substantial variation across individual manipulation tasks. We therefore analyze the complete 50-task results under Physical Random, with the full task-wise breakdown reported in Appendix~\ref{app:task-results}.

The task-wise results reveal clear structure across models. Some tasks remain consistently difficult under physical-condition diversity, while others retain relatively high success across multiple architectures. For example, manipulation tasks such as
\texttt{beat\_block\_hammer},
\texttt{move\_stapler\_pad},
\texttt{pick\_diverse\_bottles},
\texttt{put\_bottles\_dustbin},
and several stacking or placement tasks show substantially lower success for multiple models. In contrast, tasks such as
\texttt{grab\_roller},
\texttt{shake\_bottle},
\texttt{shake\_bottle\_horizontally},
and
\texttt{place\_container\_plate}
remain comparatively robust.

Several tasks consistently remain challenging across models. For example, \texttt{beat\_block\_hammer}, \texttt{pick\_diverse\_bottles}, \texttt{move\_stapler\_pad}, \texttt{place\_can\_basket}, and \texttt{put\_bottles\_dustbin} exhibit low success rates across all three WAMs. In contrast, tasks such as \texttt{grab\_roller}, \texttt{shake\_bottle}, \texttt{shake\_bottle\_horizontally}, and \texttt{place\_container\_plate} remain relatively robust.

The shared task-level structure suggests that the benchmark is not simply producing random failures. Instead, physical-condition diversity systematically stresses particular manipulation behaviors, especially those that depend on contact dynamics, object stability, mass distribution, or articulated motion.





\subsection{Discussion}
\label{sec:discussion}

Our experiments lead to two main observations. First, physical-condition diversity substantially stresses current WAMs and VLAs across a broad manipulation suite. Second, the strong agreement in task-level performance across different WAMs indicates that the challenge is structured and reproducible rather than a consequence of isolated model-specific failures.

These observations also highlight the value of the accompanying physical annotations. Once the benchmark identifies where models fail, the ground-truth physical conditions provide a basis for studying why they fail, including whether failures arise from changes in mass, friction, mass distribution, sensing configuration, or articulated dynamics.

Overall, the results support physical-condition diversity as an independent and necessary evaluation dimension for robot manipulation, and establish RoboTwin-Phys as a testbed for future work on physical-condition perception, condition-aware modeling, and adaptive robot control.

\section{Conclusion}
\label{sec:conclusion}

We present \textbf{RoboTwin-Phys}, a benchmark that introduces physical-condition diversity as an explicit dimension of robot manipulation evaluation. By continuously varying 13 physical attributes within physically plausible and task-aware domains, RoboTwin-Phys provides a systematic setting for evaluating policies under diverse physical operating conditions. The accompanying training set contains more than 5{,}000 expert demonstrations with per-episode ground-truth physical parameters, providing a reusable resource for physical-condition estimation and condition-aware policy learning.

Evaluations of representative WAMs and VLAs reveal that robustness to conventional visual and layout randomization does not necessarily extend to physical-condition diversity. The substantial performance degradation observed under physical variation exposes a distinct robustness dimension that is largely unmeasured by existing benchmarks. We hope RoboTwin-Phys can serve as a foundation for more comprehensive evaluation of embodied models and encourage future work toward models that can explicitly perceive, reason about, and adapt to changing physical conditions.

\clearpage
\beginappendix
\appendix

\section{Physical Attribute Definitions and Global Ranges}
\label{app:physical-config}

RoboTwin-Phys defines 13 physical attributes to characterize episode-level physical conditions. The attributes cover scene geometry, sensing configuration, mass, material properties, mass distribution, and joint characteristics. The complete configuration is sampled at episode initialization according to the protocol described in Sec.~\ref{sec:physical-space}.

The numerical domains used by the global configuration are listed below.

\begin{table}[h]
    \centering
    \small
    \caption{Numerical ranges of the physical attributes in the global configuration.}
    \label{tab:global-ranges}
    \begin{tabularx}{\textwidth}{@{}llX@{}}
        \toprule
        Attribute & Range & Description \\
        \midrule
        Object mass
        & $[0.125,4]\times$
        & Mass scaling, with the inertia tensor scaled accordingly. \\

        Center of mass
        & $[-1.5,1.5]$\,cm
        & Applied independently along $x/y/z$. \\

        Geometry scale
        & $[0.85,1.05]\times$
        & Uniform object scaling. \\
        \midrule

        Table tilt
        & $[0,2]^\circ$
        & Azimuth is fixed at $90^\circ$. \\

        Table height
        & $[-8,0]$\,cm
        & Relative to the nominal table height of $0.74$\,m. \\

        External force
        & magnitude $[0,0.1]\times mg$
        & Direction fixed to $+X$ and active during the policy stage. \\
        \midrule

        Camera distance \& angle
        & three modes
        & Orbital displacement $[0,10]$\,cm with gaze maintained; translation $[0,3]$\,cm without re-gazing; or rotation $[0,10]^\circ$. \\
        \midrule

        Friction $\mu$
        & $[0.01,1.25]$
        & Represents material and surface-condition differences. \\

        Restitution
        & $[0,0.5]$
        & Controls collision elasticity. \\
        \midrule

        Object joint damping
        & $[0,25]$
        & Applied to articulated objects such as drawers, cabinet doors, and switches. \\

        Arm \& gripper joint damping
        & $[0.5,3]\times$
        & Arm and gripper share one sampled scaling factor. \\
        \bottomrule
    \end{tabularx}
\end{table}

For all physical attributes, sampling is performed at the episode level and remains fixed during the corresponding rollout. The episode seed determines the random stream and therefore enables exact reproduction of the sampled condition.

\section{Task-Specific Physical Domains}
\label{app:task-config}

Although the global domains provide physically meaningful variation for most tasks, some manipulation tasks impose tighter geometric or operational constraints. Applying the global configuration to these tasks can generate combinations that are numerically valid in the simulator but inconsistent with a physically plausible instance of the task.

For example, the hammer in \texttt{beat\_block\_hammer} is relatively slender. Applying the full global CoM-offset domain can place the center of mass outside the physical body of the hammer. Such a configuration is not representative of a normal physical instance and is therefore excluded from the task-specific domain.

We consequently provide dedicated configurations for nine physics-sensitive tasks. Attributes not listed in Table~\ref{tab:per-task} retain their global settings.

\begin{table}[ht]
    \centering
    \small
    \caption{Task-specific physical configurations for the nine sensitive tasks.}
    \label{tab:per-task}
    \begin{tabular}{@{}ll@{}}
        \toprule
        Task & Dedicated adjustment \\
        \midrule
        \texttt{dump\_bin\_bigbin}
        & CoM offset narrowed to $\pm 1$\,mm \\

        \texttt{grab\_roller}
        & CoM offset narrowed to $\pm 2$\,mm \\

        \texttt{beat\_block\_hammer}
        & CoM offset narrowed to $\pm 2$\,mm \\

        \texttt{put\_bottles\_dustbin}
        & CoM offset narrowed to $\pm 2$\,mm \\

        \texttt{open\_microwave}
        & Object joint damping narrowed to $[0,0.25]$ \\

        \texttt{turn\_switch}
        & Object joint damping narrowed to $[0,1]$ \\

        \texttt{put\_object\_cabinet}
        & Size fixed at $1.0$; arm/gripper damping fixed at $1.0$ \\

        \texttt{scan\_object}
        & Camera magnitudes halved; CoM offset narrowed to $\pm 2$\,mm \\

        \texttt{place\_bread\_skillet}
        & Camera magnitudes halved; CoM offset $\pm 1$\,mm; \\
        & size $[0.9,1.05]\times$; table tilt $[0,1]^\circ$ \\
        \bottomrule
    \end{tabular}
\end{table}

The task-specific domains are selected to preserve non-trivial physical diversity while avoiding parameter combinations that are inconsistent with the geometry or operation of the corresponding task. This calibration is particularly important for factors such as center-of-mass displacement, joint damping, camera displacement, and object scaling, whose admissible ranges can vary substantially across tasks.

\section{Data Format and Organization}
\label{app:data-format}

The released demonstrations remain compatible with the official RoboTwin data format, allowing existing WAM and VLA pipelines to consume the data without modification to their trajectory interfaces.

Each task directory contains the original trajectory data and videos together with physical-condition metadata. Specifically, the release provides:

\begin{itemize}
    \item \texttt{data/}: HDF5 trajectories containing actions, states, and camera streams;
    \item \texttt{video/}: corresponding MP4 recordings;
    \item \texttt{phys\_meta/}: JSON metadata recording the 13 physical attributes instantiated for each episode;
\end{itemize}

A dataset manifest further records the mapping between tasks, benchmark configurations, and episode counts.

\section{Per-Task Physical Random Results}
\label{app:task-results}

Table~\ref{tab:per-task-phys} reports the complete task-level success rates of the five evaluated models under the RoboTwin-Phys \texttt{phys\_random\_all} setting. Each task contains 100 rollout episodes.

\begin{table*}[t]
    \centering
    \small
    \caption{Per-task success rates (\%) under the RoboTwin-Phys Physical Random setting. Each task is evaluated with 100 rollout episodes.}
    \label{tab:per-task-phys}
    \begin{tabular}{@{}lrrrrr@{}}
        \toprule
        Task & Fast-WAM & Motus & FACT & $\pi_{0.5}$ & Galaxea-VLA \\
        \midrule
        \texttt{adjust\_bottle} & 64 & 39 & 43 & 43 & 65 \\
        \texttt{beat\_block\_hammer} & 17 & 13 & 11 & 8 & 8 \\
        \texttt{blocks\_ranking\_rgb} & 57 & 46 & 48 & 38 & 36 \\
        \texttt{blocks\_ranking\_size} & 56 & 35 & 21 & 13 & 44 \\
        \texttt{click\_alarmclock} & 69 & 61 & 36 & 34 & 59 \\
        \texttt{click\_bell} & 57 & 62 & 23 & 46 & 36 \\
        \texttt{dump\_bin\_bigbin} & 32 & 34 & 33 & 65 & 24 \\
        \texttt{grab\_roller} & 65 & 72 & 77 & 91 & 72 \\
        \texttt{handover\_block} & 33 & 18 & 18 & 11 & 23 \\
        \texttt{handover\_mic} & 54 & 37 & 51 & 15 & 43 \\
        \texttt{hanging\_mug} & 26 & 15 & 9 & 17 & 15 \\
        \texttt{lift\_pot} & 29 & 33 & 34 & 29 & 30 \\
        \texttt{move\_can\_pot} & 50 & 26 & 50 & 2 & 34 \\
        \texttt{move\_playingcard\_away} & 59 & 61 & 57 & 46 & 43 \\
        \texttt{move\_stapler\_pad} & 17 & 24 & 15 & 0 & 23 \\
        \texttt{open\_laptop} & 59 & 49 & 64 & 45 & 51 \\
        \texttt{open\_microwave} & 17 & 42 & 48 & 17 & 33 \\
        \texttt{pick\_diverse\_bottles} & 17 & 19 & 13 & 19 & 27 \\
        \texttt{pick\_dual\_bottles} & 33 & 21 & 32 & 14 & 35 \\
        \texttt{place\_a2b\_left} & 54 & 49 & 48 & 33 & 51 \\
        \texttt{place\_a2b\_right} & 49 & 44 & 50 & 18 & 57 \\
        \texttt{place\_bread\_basket} & 50 & 48 & 39 & 38 & 50 \\
        \texttt{place\_bread\_skillet} & 39 & 44 & 48 & 31 & 44 \\
        \texttt{place\_burger\_fries} & 53 & 49 & 49 & 74 & 38 \\
        \texttt{place\_can\_basket} & 17 & 19 & 19 & 15 & 20 \\
        \texttt{place\_cans\_plasticbox} & 28 & 15 & 34 & 27 & 15 \\
        \texttt{place\_container\_plate} & 66 & 61 & 65 & 23 & 48 \\
        \texttt{place\_dual\_shoes} & 28 & 25 & 22 & 41 & 23 \\
        \texttt{place\_empty\_cup} & 57 & 50 & 50 & 39 & 35 \\
        \texttt{place\_fan} & 38 & 28 & 36 & 16 & 32 \\
        \texttt{place\_burger\_fries} & 53 & 49 & 49 & 74 & 38 \\
        \texttt{place\_mouse\_pad} & 35 & 17 & 22 & 22 & 22 \\
        \texttt{place\_object\_basket} & 41 & 43 & 35 & 2 & 31 \\
        \texttt{place\_object\_scale} & 46 & 39 & 46 & 6 & 34 \\
        \texttt{place\_object\_stand} & 53 & 52 & 51 & 49 & 40 \\
        \texttt{place\_phone\_stand} & 52 & 55 & 49 & 45 & 37 \\
        \texttt{move\_pillbottle\_pad} & 43 & 31 & 32 & 2 & 26 \\
        \texttt{place\_shoe} & 63 & 60 & 58 & 25 & 44 \\
        \texttt{press\_stapler} & 58 & 72 & 46 & 38 & 42 \\
        \texttt{put\_bottles\_dustbin} & 11 & 9 & 14 & 25 & 38 \\
        \texttt{put\_object\_cabinet} & 39 & 12 & 32 & 18 & 42 \\
        \texttt{rotate\_qrcode} & 43 & 29 & 42 & 31 & 33 \\
        \texttt{scan\_object} & 29 & 18 & 44 & 63 & 39 \\
        \texttt{shake\_bottle} & 69 & 68 & 68 & 87 & 71 \\
        \texttt{shake\_bottle\_horizontally} & 75 & 69 & 71 & 75 & 76 \\
        \texttt{stack\_blocks\_three} & 30 & 25 & 10 & 14 & 29 \\
        \texttt{stack\_blocks\_two} & 63 & 56 & 50 & 11 & 50 \\
        \texttt{stack\_bowls\_three} & 52 & 38 & 36 & 39 & 32 \\
        \texttt{stack\_bowls\_two} & 62 & 60 & 56 & 65 & 36 \\
        \texttt{stamp\_seal} & 24 & 23 & 16 & 12 & 15 \\
        \texttt{turn\_switch} & 34 & 63 & 36 & 38 & 40 \\
        \midrule
        Average & 44.24 & 39.60 & 39.14 & 31.60 & 37.82 \\
        \bottomrule
    \end{tabular}
\end{table*}

\clearpage
\bibliographystyle{plainnat}
\bibliography{ref}

@inproceedings{bi2025motus,
      title={Motus: A Unified Latent Action World Model}, 
      author={Hongzhe Bi and Hengkai Tan and Shenghao Xie and Zeyuan Wang and Shuhe Huang and Haitian Liu and Ruowen Zhao and Yao Feng and Chendong Xiang and Yinze Rong and Hongyan Zhao and Hanyu Liu and Zhizhong Su and Lei Ma and Hang Su and Jun Zhu},
      year={2026},
      booktitle={Conference on Computer Vision and Pattern Recognition (CVPR)},
      pages={35101-35113}
}

@inproceedings{yuan2026fastwam,
      title={Fast-WAM: Do World Action Models Need Test-time Future Imagination?}, 
      author={Tianyuan Yuan and Zibin Dong and Yicheng Liu and Hang Zhao},
      year={2026},
      booktitle={arXiv preprint arXiv:2603.16666},
}

@inproceedings{peng2026fact,
  title     = {FACT: Failure-Aware Causal Training for World-Action Models},
  author    = {Peng, Quanquan and Liang, Yutong and Yan, Rui and Hansen, Nicklas and Wang, Xiaolong},
  booktitle = {Conference on Robot Learning (CoRL)},
  year      = {2026}
}

@inproceedings{chen2025robotwin,
        title={RoboTwin 2.0: A Scalable Data Generator and Benchmark with Strong Domain Randomization for Robust Bimanual Robotic Manipulation},
        author={Chen, Tianxing and Chen, Zanxin and Chen, Baijun and Cai, Zijian and Liu, Yibin and Liang, Qiwei and Li, Zixuan and Lin, Xianliang and Ge, Yiheng and Gu, Zhenyu and others},
        year={2025},
        booktitle={arXiv preprint arXiv:2506.18088},
      }

@inproceedings{intelligence2025pi05,
      title={$\pi_{0.5}$: a Vision-Language-Action Model with Open-World Generalization}, 
      author={Physical Intelligence and Kevin Black and Noah Brown and James Darpinian and Karan Dhabalia and Danny Driess and Adnan Esmail and Michael Equi and Chelsea Finn and Niccolo Fusai and Manuel Y. Galliker and Dibya Ghosh and Lachy Groom and Karol Hausman and Brian Ichter and Szymon Jakubczak and Tim Jones and Liyiming Ke and Devin LeBlanc and Sergey Levine and Adrian Li-Bell and Mohith Mothukuri and Suraj Nair and Karl Pertsch and Allen Z. Ren and Lucy Xiaoyang Shi and Laura Smith and Jost Tobias Springenberg and Kyle Stachowicz and James Tanner and Quan Vuong and Homer Walke and Anna Walling and Haohuan Wang and Lili Yu and Ury Zhilinsky},
      year={2025},
        booktitle={arXiv preprint arXiv:2504.16054},
}

@inproceedings{liu2026g05auto,
  title={G0.5: One Autoregressive Stream for Robot Reasoning and Action},
  author={Yicheng Liu and Zibin Dong and Baijun Ye and Tianyuan Yuan and Tao Jiang and Anqi Yang and Shicheng Cao and Haonan Liu and Yue Sun and Zihan Guo and Xiao Liu and Ke Dong and Changxun Pan and Chenru Wu and Tailai Cheng and Xiaoshu Ren and Xinlei Zhang and Jianning Cui and Zijie Zhao and Haoyu Zhang and Kaiming Xu and Haodong Yang and Bowen Zhang and Jiahui Niu and Shaoting Zhu and Shiduo Zhang and Hang Zhao},
  year={2026},
  booktitle={arXiv preprint arXiv:2608.11739},
}

\end{document}